\documentclass[sigconf]{acmart}

\usepackage{float, graphicx}
\usepackage{lipsum}
\usepackage{multirow}
\usepackage{tabularx}
\usepackage{csquotes}
\usepackage{dblfloatfix}
\AtBeginDocument{%
  }

\usepackage{bm}
\usepackage{amsmath,scalerel}
\usepackage{xfrac}
\usepackage{bbm}
\usepackage{graphicx}
\usepackage{subfig}
\usepackage{algpseudocode}
\usepackage{algorithm}

\begin{document}



\title{Real-Time Inverse Kinematics for Generating Multi-Constrained Movements of Virtual Human Characters}



\author{Hendric Voss}
\email{hvoss@techfak.uni-bielefeld.de}
\affiliation{%
  \institution{Social Cognitive Systems Group}
  \institution{Bielefeld University}
  \streetaddress{Universitätsstraße 25}
  \country{Germany}
}

\author{Stefan Kopp}
\email{skopp@techfak.uni-bielefeld.de}
\affiliation{%
  \institution{Social Cognitive Systems Group}
  \institution{Bielefeld University}
  \streetaddress{Universitätsstraße 25}
  \country{Germany}
}

\renewcommand{\shortauthors}{Voss et al.}

\begin{abstract}
Generating accurate and realistic virtual human movements in real-time is of high importance for a variety of applications in computer graphics, interactive virtual environments, robotics, and biomechanics. This paper introduces a novel real-time inverse kinematics (IK) solver specifically designed for realistic human-like movement generation. Leveraging the automatic differentiation and just-in-time compilation of TensorFlow, the proposed solver efficiently handles complex articulated human skeletons with high degrees of freedom. By treating forward and inverse kinematics as differentiable operations, our method effectively addresses common challenges such as error accumulation and complicated joint limits in multi-constrained problems, which are critical for realistic human motion modeling. We demonstrate the solver's effectiveness on the SMPLX human skeleton model, evaluating its performance against widely used iterative-based IK algorithms, like Cyclic Coordinate Descent (CCD), FABRIK, and the nonlinear optimization algorithm IPOPT. Our experiments cover both simple end-effector tasks and sophisticated, multi-constrained problems with realistic joint limits. Results indicate that our IK solver achieves real-time performance, exhibiting rapid convergence, minimal computational overhead per iteration, and improved success rates compared to existing methods. The project code is available at \url{https://github.com/hvoss-techfak/JAX-IK}

\end{abstract}




\maketitle


\section{Introduction}
Modern robotics, computer graphics, and biomechanics often face the challenge of moving complex articulated structures, ranging from robotic arms to digital human characters, to achieve desired poses or movements \cite{keller2023skin,pechev2008inverse}. Accurate and efficient motion generation requires solving the inverse kinematics (IK) problem: determining a set of joint configurations that achieve a desired position or orientation of an end-effector \cite{murray2017mathematical}. In many applications, from animating lifelike characters in virtual environments to controlling multi-joint robotic manipulators, IK provides the essential computational framework for translating high-level motion goals into precise joint movements \cite{murray2017mathematical}.

Solving the IK problem is particularly acute in the generation of real-time, realistic movements for virtual human characters, where systems must simultaneously address speed, accuracy, and the physical and anatomical plausibility of human-like motion \cite{ jiang2024manikin}. This is inherently challenging due to the complexity of the human skeleton, with its intricate assembly of joints and highly non-linear and interdependent degrees of freedom \cite{liu_computational_2021}. 
In the field of humanoid robotics and virtual agents, this challenge is compounded, as systems must not only mimic the natural motion of the human body but also operate in real-time to enable seamless interaction and engagement \cite{montecillo-puente_real-time_2010,diomataris_wandr_2024,sakka_tasks_2014}.

Traditional IK approaches, such as Jacobian-based iterative methods \cite{duleba_comparison_2013}, often struggle with local minima or require significant simplifications that compromise the naturalness of motion \cite{peiper_kinematics_nodate}. Furthermore, in a high degree of freedom system, even small numerical errors can propagate along the kinematic chain, leading to large discrepancies at the end effectors \cite{duleba_comparison_2013,canutescu_cyclic_2003,aristidou2011fabrik}. This is especially critical when modeling human-like motion under multi-constraint conditions, where arbitrary and dynamic constraints, such as joint limits, collision avoidance, contact interactions, and expressive behavior, must be accommodated simultaneously \cite{sui2025survey}, sometimes even in dynamic environments and real-time interactive applications involving human characters. These challenges highlight the need for IK solvers that can efficiently handle complex kinematic chains with multiple constraints while maintaining both accuracy and flexibility.

In this paper, we present a TensorFlow-based IK solver designed for the real-time generation of multi-constrained movements in virtual human characters. Emphasizing simplicity, speed, and accuracy, our approach leverages the automatic differentiation and just-in-time (JIT) compilation features of TensorFlow \cite{tensorflow2015-whitepaper} to formulate both forward and inverse kinematics as fully differentiable functions. This design not only accelerates the computation for complex kinematic chains but also allows the creation of arbitrarily complex objective functions that can be used to model any number of different end-effector constraints. Our solver is particularly well suited for a high degree of freedom human body models, where minimizing the error between target positions and computed joint configurations is challenging due to complex rotational dynamics and strict boundary conditions along the kinematic tree.

\section{Related Work}

As one of the earliest approaches to inverse kinematics, analytical methods solve a set of closed-form equations to position an end effector at a specified target. While effective, these methods have historically relied on strong assumptions about a robot's specific structure \cite{peiper_kinematics_nodate}. More recently, analytical solutions have emerged that are independent of predefined structural constraints. \citet{tian_analytical_2021} proposed a solution for 7-DOF anthropomorphic manipulators by decomposing the problem into eight groups of arm configuration manifolds.

In contrast, numerical methods use iterative algorithms that converge to a solution based on an initial estimate. A fundamental numerical approach is the Moore-Penrose pseudo-inverse method, introduced by \citet{barata_moorepenrose_2012}, which determines joint configurations by minimizing the least-squares error between the current and desired end-effector positions. To improve stability, especially near singular configurations, \citet{buss_selectively_2005} introduced the Selectively Damped Least Squares (SDLS) method, which dynamically adjusts the damping factor to mitigate instability.

Another widely used numerical technique is Cyclic Coordinate Descent (CCD) \cite{canutescu_cyclic_2003}, which is particularly useful in applications that require computational efficiency, such as real-time animation and robotics. CCD iteratively adjusts each joint in a kinematic chain, working backward from the end effector to the base to progressively minimize positional error.

Forward And Backward Reaching Inverse Kinematics (FABRIK) \cite{aristidou2011fabrik} refines joint positions by successively adjusting them in forward and backward passes along the kinematic chain. Unlike CCD, which rotates joints to minimize end-effector error, FABRIK directly manipulates joint positions while maintaining link constraints, resulting in smoother and more stable solutions. This makes it ideal for character animation and humanoid robotics applications where natural motion is essential.

Similarly, Interior Point OPTimizer (IPOPT) \cite{wachter2002interior,wachter2006implementation} is a primal-dual interior-point solver for large-scale nonlinear programming. It tackles a sequence of logarithmic-barrier sub-problems and uses a filter line-search to separately enforce decreases in constraint violation or objective value, which guarantees global convergence without resorting to merit functions. Each Newton step is complemented by second-order corrections, inertia-controlled linear solves and a dedicated feasibility-restoration phase, giving the method robustness even from poor initial guesses. This approach makes IPOPT a preferred choice for trajectory optimisation, process engineering and other engineering optimisation tasks where both scalability and numerical reliability are critical.

Further advances in real-time motion synthesis were made by \citet{rakita2018relaxedik}, who developed a method that ensures smooth, feasible manipulator motions while avoiding joint space discontinuities by framing inverse kinematics as a weighted sum nonlinear optimization problem. Similarly, \citet{zhang2024real} proposed an approach based on constrained linear programming that incorporates joint bounds and velocity and acceleration bounds as inequality constraints to refine kinematic solutions. In addition, nonlinear solvers such as Sequential Quadratic Programming (SQP) \cite{boggs1995sequential} and Interior-Point Methods (IPM) \cite{nemirovski_interior-point_2008} have been used to efficiently solve multi-constraint inverse kinematic problems.

\subsection{Machine Learning Inverse Kinematic}
Another important development in numerical inverse kinematics is the use of machine learning approaches to learn optimization-based techniques. Current implementations of inverse kinematics in the machine learning domain focus mainly on training neural networks to learn the IK problem from training data, and using these trained models during the inference phase to solve the IK problem for new, unseen end-effector targets \cite{bensadoun2022neural}. As one of the earliest approaches, \citet{koker_study_2004} trained an artificial neural network (ANN) on five thousand trajectories and inferred cubic trajectories directly from the neural network output. 

Building on this foundation, more recent advances have used deep learning architectures to improve the generalization and efficiency of inverse kinematics (IK) solutions. \citet{levine_end-end_2016} demonstrates the potential of deep reinforcement learning (RL) to solve IK problems in robotic manipulation tasks without requiring explicit kinematic modeling. By training policies to maximize task performance, RL-based IK methods can adapt to different environments and constraints.

A notable extension of this paradigm is the MAPPO IK algorithm proposed by \citet{zhao_inverse_2024}, which uses multi-agent proximal policy optimization to simultaneously optimize position and posture control in six-degree-of-freedom manipulators. Unlike previous RL implementations that focus primarily on positional accuracy, this approach introduces a dual reward mechanism that combines Gaussian distance for positional accuracy and cosine distance for orientation alignment.

The challenge of solution multiplicity in redundant manipulators has inspired alternative machine learning frameworks. \citet{bocsi_learning_2011} pioneered structured output learning using joint kernel support vector machines to model the probability distribution of end-effector positions and joint angles. By embedding sine/cosine joint angle transformations in kernel space, their method achieved 1.4 mm tracking accuracy on a 7-DOF Barrett WAM while maintaining temporal consistency across ambiguous configurations.

Hybrid analytic-statistical approaches have emerged to bridge the gap between model-based and data-driven methods. \citet{nguyen_modeling_2022} developed a Gaussian process framework that uses closed-form kinematic equations as prior distributions updated by Bayesian inference with optical motion capture data, significantly reducing calibration points compared to pure neural network approaches.

Recent innovations in neural architectures address inverse kinematics through hierarchical probability modeling. The IKNet framework by \citet{bensadoun_neural_2022} uses variational autoencoder principles with Gaussian mixture models to sequentially sample joint configurations along the kinematic chain. By conditioning the probability distribution of each joint on previous angles and end-effector targets, the method generates multiple physically plausible solutions with high success rates.

Although powerful, \citet{Eapen2023ComparativeSO} showed that Gaussian regression and support vector machine algorithms often outperform neural network approaches for IK solutions.

\section{Architecture and Method}

In this section, we describe the architecture and methodology of our IK solver. Our system consists of two main components: a pure, JIT-compiled forward kinematics (FK) module and an inverse kinematics solver that minimizes an objective function formulated from the IK error and additional biomechanical penalties. We first outline the forward kinematics formulation, then describe the inverse kinematics optimization framework, discuss the integration of TensorFlow for automatic differentiation and gradient descent, and finally explain the various objective functions used to enforce kinematic and task-specific constraints.

\subsection{Forward Kinematics}

The FK module forms the basis of the solver by computing global transformations for a given human skeleton. A skeleton is loaded and represented as a tree structure with nodes corresponding to bones. A local transformation matrix is defined for each bone, and the FK computation uses these local transformations along with parent-child relationships to compute the global pose. 

To update controlled bones, the forward kinematics (FK) model computes the global transformation of each joint in the kinematic tree from the local joint transformations. For a joint \(i\) with parent index \(p(i)\), its transformation \(T_i\) is recursively computed as
\begin{equation}
    T_i = 
    \begin{cases}
      L_i R_i, & \text{if } p(i) < 0 \\
      T_{p(i)} \, L_i R_i, & \text{otherwise}
    \end{cases}
    \label{eq:fk_recursion}
\end{equation}
where \(L_i \in \mathbb{R}^{4\times4}\) is the fixed local transformation (encoding the length and rest pose of the bone), and \(R_i \in \mathbb{R}^{4\times4}\) is the rotation matrix associated with the joint \(i\).

The rotation matrix \(R_i\) is derived from the Euler angles \(\boldsymbol{\theta} = [\theta_x, \theta_y, \theta_z]^\top\). Specifically, we compute the rotation matrices about the \(x\)-, \(y\)-, and \(z\)-axes as follows:
\begin{align}
    R_x(\theta_x) &= \begin{bmatrix}
    1 & 0 & 0 & 0 \\
    0 & \cos\theta_x & -\sin\theta_x & 0 \\
    0 & \sin\theta_x & \cos\theta_x & 0 \\
    0 & 0 & 0 & 1
    \end{bmatrix}, \label{eq:Rx}\\[1ex]
    R_y(\theta_y) &= \begin{bmatrix}
    \cos\theta_y & 0 & \sin\theta_y & 0 \\
    0 & 1 & 0 & 0 \\
    -\sin\theta_y & 0 & \cos\theta_y & 0 \\
    0 & 0 & 0 & 1
    \end{bmatrix}, \label{eq:Ry}\\[1ex]
    R_z(\theta_z) &= \begin{bmatrix}
    \cos\theta_z & -\sin\theta_z & 0 & 0 \\
    \sin\theta_z & \cos\theta_z & 0 & 0 \\
    0 & 0 & 1 & 0 \\
    0 & 0 & 0 & 1
    \end{bmatrix}. \label{eq:Rz}
\end{align}
The combined rotation is then given by:
\begin{equation}
    R_i = R_z(\theta_z) \, R_y(\theta_y) \, R_x(\theta_x).
    \label{eq:combined_rotation}
\end{equation}

The FK is computed over the entire chain, resulting in a set of global transformations \(\{T_i\}\) that are used for further IK computations. This differentiable and modular FK pipeline ensures that gradients can be propagated back through the kinematic chain, a key requirement for the subsequent IK optimization.

\subsection{Inverse Kinematics}

The inverse kinematics (IK) problem seeks the optimal set of joint angles \(\boldsymbol{\theta}\) that minimizes a given error function \(J(\boldsymbol{\theta})\). An iterative gradient-descent-based optimization routine refines the angle vector over several iterations. At each iteration, the gradient of the objective function is computed with respect to the joint angles, and a learning rate is applied to update the solution while respecting the lower and upper bounds of the joint angles. 

In our formulation, the objective function is constructed as:
\begin{equation}
    J(\boldsymbol{\theta}) = J_{\text{main}}(\boldsymbol{\theta}) + \sum_{k} J_{k}(\boldsymbol{\theta}),
    \label{eq:total_objective}
\end{equation}
where \(J_{\text{main}}\) is the primary objective (typically enforcing the distance between the end-effector and the target) and the additional terms \(\{J_{k}\}\) incorporate further constraints such as orientation alignment or collision avoidance.

\subsection{TensorFlow Integration and Gradient Descent}

Our implementation uses the TensorFlow framework to provide efficient automatic differentiation and just-in-time (JIT) compilation \cite{tensorflow2015-whitepaper}. By expressing all kinematic computations as TensorFlow functions, we can compute gradients of the objective function \(J(\boldsymbol{\theta})\) with respect to the joint angles \(\boldsymbol{\theta}\). In machine learning tasks, the gradient descent algorithm updates the joint angles by computing the gradient of the objective function and then applying the update rule:
\begin{equation}
    \boldsymbol{\theta}_{t+1} = \boldsymbol{\theta}_t - \eta \, \nabla J(\boldsymbol{\theta}_t),
    \label{eq:vanilla_gd}
\end{equation}
where \(\eta\) is the learning rate.

To improve convergence and stability, we use the Adam gradient descent algorithm \cite{kingma_adam_2017}. Adam refines the gradient descent update rule by maintaining exponential moving averages of the gradients and their squares. At each iteration \(t\), these moment estimates are computed as
\begin{align}
    m_{t+1} &= \beta_1 m_t + (1-\beta_1) \nabla J(\boldsymbol{\theta}_t), \label{eq:adam_m}\\[1ex]
    v_{t+1} &= \beta_2 v_t + (1-\beta_2) \left(\nabla J(\boldsymbol{\theta}_t)\right)^2, \label{eq:adam_v}
\end{align}
with bias-corrected estimates given by:
\begin{align}
    \hat{m}_{t+1} &= \frac{m_{t+1}}{1-\beta_1^{t+1}}, \label{eq:adam_m_hat}\\[1ex]
    \hat{v}_{t+1} &= \frac{v_{t+1}}{1-\beta_2^{t+1}}. \label{eq:adam_v_hat}
\end{align}
The joint angles are then updated according to:
\begin{equation}
    \boldsymbol{\theta}_{t+1} = \boldsymbol{\theta}_t - \eta \frac{\hat{m}_{t+1}}{\sqrt{\hat{v}_{t+1}} + \epsilon},
    \label{eq:adam_update}
\end{equation}
where \(\epsilon\) is a small constant for numerical stability.

In addition to the standard Adam update, we incorporate a cautious modification to suppress unstable updates. Inspired by recent work on cautious optimizers \cite{liang2024cautious}, we compute a binary mask \(M\) that retains only those coordinates where the bias-corrected momentum \(\hat{m}_t\) and the gradient \(\nabla J(\boldsymbol{\theta}_t)\) share the same sign, i.e.,
\begin{equation}
M = \frac{\mathbb{I}\Bigl(\hat{m}_t \circ \nabla J(\boldsymbol{\theta}_t) > 0\Bigr)}{\max\Bigl(\mathrm{mean}\Bigl(\mathbb{I}\Bigl(\hat{m}_t \circ \nabla J(\boldsymbol{\theta}_t) > 0\Bigr)\Bigr),\, \epsilon\Bigr)},
\end{equation}
where \(\circ\) denotes the element-wise product and \(\epsilon\) is a small constant for numerical stability. This mask is then applied to weight the momentum update, zeroing out contributions with inconsistent sign information. A compensatory scaling factor,
\begin{equation}
\alpha(x) = \frac{\mathrm{dim}(x)}{\mathrm{nnz}(x>0)+1},
\end{equation}
ensures that the overall update magnitude is preserved. Consequently, the cautious update
\begin{equation}
\tilde{m}_t = \alpha(\hat{m}_t)\Bigl(\hat{m}_t \circ M\Bigr)
\end{equation}
is used in place of \(\hat{m}_t\) in the parameter update rule, leading to enhanced convergence and stability during training \cite{liang2024cautious}.

\subsection{Objective Functions}

The design of the objective functions is critical to ensure both accuracy and natural motion. We implement several objective functions in our framework:

\paragraph{Distance Objective}
\label{distance_obj}
This term penalizes the Euclidean distance between the computed end-effector position and the target:
\begin{equation}
    J_{\text{distance}}(\boldsymbol{\theta}) = \left\| \mathbf{t} - T_{\text{effector}}(\boldsymbol{\theta}) \right\|_2.
    \label{eq:distance_obj_repeat}
\end{equation}

\paragraph{Look-At Objective}  
To enforce a desired orientation, such as aligning a bone with a target direction, we define a look-at penalty:
\begin{equation}
    J_{\text{look}}(\boldsymbol{\theta}) = \left( \arccos \frac{\langle \mathbf{d}_b(\boldsymbol{\theta}), \mathbf{d}_t \rangle}{\|\mathbf{d}_b(\boldsymbol{\theta})\|\|\mathbf{d}_t\|} \right)^2,
    \label{eq:look_obj}
\end{equation}
where \(\mathbf{d}_b(\boldsymbol{\theta})\) is the direction of the bone computed from the FK, and \(\mathbf{d}_t\) is the target direction. The formulation involves modifications to the target point, enabling fine-tuning of the look-at behavior.

\paragraph{Known Rotation Objective}  
For scenarios where a particular joint rotation is desired, we introduce a penalty that measures the deviation from a candidate rotation \(\boldsymbol{\theta}^*\):
\begin{equation}
    J_{\text{known}}(\boldsymbol{\theta}) = \frac{1}{N} \sum_{i=1}^{N} \left\| \theta_i - \theta_i^* \right\|_2^2,
    \label{eq:known_obj}
\end{equation}
with \(N\) being the number of controlled degrees of freedom. A binary mask can be applied to focus the penalty on specific joints.

\paragraph{Objective Combination}
The overall IK optimization can performed by composing any of the above objectives. For example, for our evaluation, we combined all equations:
\begin{equation}
    \begin{aligned}
        J(\boldsymbol{\theta}) = &\ \lambda_{\text{dist}} J_{\text{distance}}(\boldsymbol{\theta}) 
        + \lambda_{\text{look}} J_{\text{look}}(\boldsymbol{\theta}) \\
        &\ + \lambda_{\text{known}} J_{\text{known}}(\boldsymbol{\theta}).
    \end{aligned}
    \label{eq:full_objective}
\end{equation}
where \(\lambda_{\text{dist}}, \lambda_{\text{look}}, \lambda_{\text{known}}\) are weights that balance the contribution of each objective.

The optimization process leverages the forward kinematics module to compute the global transformations, and the gradient descent module iteratively updates \(\boldsymbol{\theta}\) to minimize \(J(\boldsymbol{\theta})\) subject to bound constraints:
\begin{equation}
    \boldsymbol{\theta} \in [\boldsymbol{\theta}_{\min}, \boldsymbol{\theta}_{\max}].
    \label{eq:bounds}
\end{equation}

\begin{figure*}[tbh]
    \centering
    \includegraphics[width=0.9\linewidth]{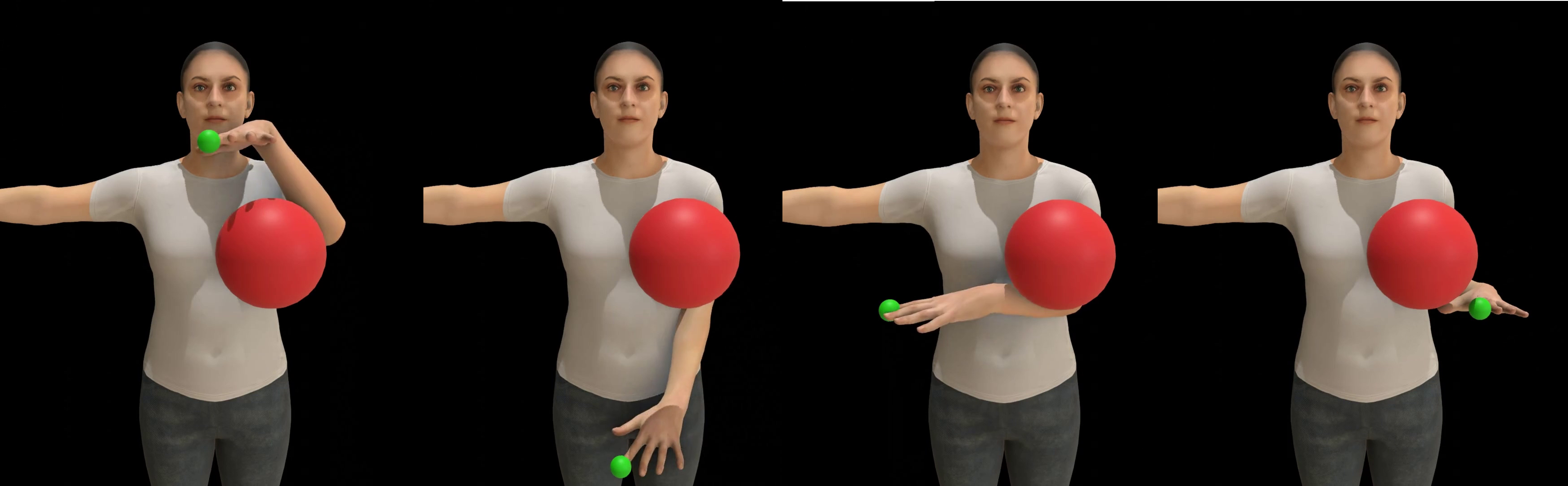}
    \caption{Visualization of multiple combined objectives. The green ball is the target position, while the red ball has to be avoided. We combine the \textit{Distance Objective}, a hand palm down \textit{Look-At Objective}, an optional hand forward \textit{Look-At Objective} and the motion planning \textit{Smoothness Objective}. Full demo videos are available at \url{https://hvoss-techfak.github.io/TF-JAX-IK/}}\label{fig:visualization}
\end{figure*}

\subsection{Extendability}

This modular design allows for the easy integration of additional objectives and constraints, making the solver highly adaptable for various applications in the field of robotics and virtual agents. The following pseudo-code shows, on an example of the distance objective described in section \ref{distance_obj}, the ease of implementing new objective functions for the IK system. Through the use of JIT compiling and static arguments, arbitrarily complex functions can be created. 

\begin{algorithm}
\caption{Distance Objective Function}
\begin{algorithmic}[1]
\Require $\text{fksolver}$, $\text{target\_point}$, $\text{bone\_name}$
\Function{distance\_obj}{$\text{fksolver}, \text{target\_point}, \text{bone\_name}$}
    \State $\text{fk\_bone} \gets \text{fksolver}(\text{bone\_name})$
    \State \Return $\|\text{fk\_bone} - \text{target\_point}\|$
\EndFunction
\end{algorithmic}
\label{objective_example}
\end{algorithm}

\subsection{Stopping Criteria}

The optimization loop in our IK solver incorporates three key stopping criteria to ensure both convergence and real-time performance. Let \( t \) denote the current iteration, \( N_{\max} \) the user-defined maximum number of iterations, and \( J(\boldsymbol{\theta}_t) \) the objective function value at iteration \( t \).

\paragraph{Iteration Count Criterion}  
The first criterion ensures that the solver does not run indefinitely by terminating the optimization when the number of iterations exceeds a predefined maximum:
\begin{equation}
    t \geq N_{\max}.
    \label{eq:iter_stop}
\end{equation}

\paragraph{Error Threshold Criterion}  
The second criterion halts the optimization once the objective function (i.e., the error) falls below a specified threshold \(\epsilon\):
\begin{equation}
    J(\boldsymbol{\theta}_t) < \epsilon.
    \label{eq:error_stop}
\end{equation}

\paragraph{Dynamic Iteration Adjustment}  
In addition to the fixed iteration and error criteria, the solver dynamically adjusts the maximum allowed iterations based on the average computation time per iteration. Let \( T_{\max} \) represent the maximum allowed total computation time and \(\bar{t}\) the average time per iteration:
\begin{equation}
    \bar{t} = \frac{1}{t} \sum_{i=1}^{t} t_i,
    \label{eq:avg_iter_time}
\end{equation}
where \( t_i \) is the computation time for the \( i \)-th iteration. The maximum allowed iterations, \( N_{\text{allowed}} \), are then determined as:
\begin{equation}
    N_{\text{allowed}} = \min \left\{ N_{\max}, \left\lfloor \frac{T_{\max}}{\bar{t}} \right\rfloor \right\}.
    \label{eq:dynamic_iter}
\end{equation}

Collectively, these stopping criteria ensure that the optimization process terminates either when a satisfactory solution is reached (i.e. when the error is sufficiently low), the iteration limit is met, or the time constraints are exceeded, thereby balancing computational efficiency and solution accuracy.

\subsection{Motion Planning}

Our IK solver can be naturally extended to perform motion planning, finding smooth transitions between an initial pose and a target configuration through intermediate points. To achieve this, we introduce a trajectory consisting of the original position, a predefined number of intermediate subpoints, and the final target position. The goal is to optimize the entire trajectory simultaneously, resulting in natural and smooth motions.

To ensure smoothness throughout the planned trajectory, we generalize the smoothness objectives using finite differences. The generalized smoothness objective penalizes rapid changes in the trajectory of joint angles, formulated as:
\begin{equation}
    J_{\text{smoothness}}^{(n)}(\boldsymbol{\theta}) = \sum_{t=1}^{T-n} \left\| \sum_{k=0}^{n} (-1)^k \binom{n}{k} \boldsymbol{\theta}_{t+n-k} \right\|_2^2,
    \label{eq:smoothness_general}
\end{equation}
where \( n \) represents the order of the finite difference (velocity \( n=1 \), acceleration \( n=2 \), jerk \( n=3 \)), and \( T \) is the total number of points in the trajectory.

The optimization framework combines this generalized smoothness objective with previously defined task-specific objectives (e.g., distance, look-at, known rotations, collision avoidance) to construct the complete motion-planning objective:
\begin{equation}
\begin{aligned}
J_{\text{trajectory}}(\boldsymbol{\theta}) &= 
 \lambda_{\text{dist}} J_{\text{distance}}(\boldsymbol{\theta})
+ \lambda_{\text{look}} J_{\text{look}}(\boldsymbol{\theta}) \\
&+ \lambda_{\text{known}} J_{\text{known}}(\boldsymbol{\theta}) \\
&+ \sum_{n=1}^{3} \lambda_n J_{\text{smoothness}}^{(n)}(\boldsymbol{\theta}),
\end{aligned}
\label{eq:full_trajectory_objective}
\end{equation}
where the weights \( \lambda_{\text{dist}}, \lambda_{\text{look}}, \lambda_{\text{known}}, \lambda_n \) balance the influence of each term in the overall optimization.

The solver initializes a trajectory vector \( \boldsymbol{\theta} \) containing the joint angles for all points (initial, intermediate, and target). The Adam optimizer iteratively refines this trajectory by minimizing \( J_{\text{trajectory}}(\boldsymbol{\theta}) \) while adhering to joint-angle constraints:
\begin{equation}
    \boldsymbol{\theta}_t \in [\boldsymbol{\theta}_{\min}, \boldsymbol{\theta}_{\max}], \quad \text{for all trajectory points } t.
    \label{eq:trajectory_bounds}
\end{equation}

This approach yields a smooth and natural motion sequence, with a variable amount of intermediate points, from the initial configuration to the final target, while adhering to all given objective functions. Figure \ref{fig:visualization} shows a combination of multiple objectives together with the smoothness objective.


\section{Experimental Results}

\begin{table}[bt]
\begin{tabular}{ccc}
\hline
Implementation & Device & Avg. Time per Iteration (ms) \\ 
\hline
JAX & GPU & 1.2 $\pm$ 4.0 \\
JAX & CPU & 0.5 $\pm$ 1.6 \\
TensorFlow & GPU & 3.9 $\pm$ 13.7 \\
TensorFlow & CPU & \textbf{0.3 $\pm$ 0.8} \\
\hline

\end{tabular}
\caption{Comparison of iteration times between the CPU and GPU configurations of both the TensorFlow and JAX implementation. In all fields we report the mean value with the respective standard deviation. }
\label{tab:cpu_gpu}
\end{table}

\begin{table*}[hbt]
\centering
\begin{tabular}{c c c c c c}
\hline
Algorithm & Custom Objective & Solving Time (ms) & Iterations & Time per Iteration (ms) & Success Rate (\%) \\
\hline
CCD & No & 25.44 $\pm$ 62.61 & 53.57 $\pm$ 139.19 & 0.69 $\pm$ 0.27 & 91.19 \\
FABRIK & No & 85.28 $\pm$ 26.92 & 453.38 $\pm$ 144.18 & 0.23 $\pm$ 0.16 & 96.12 \\
IPOPT & No & 164.73 $\pm$ 132.25 & \textbf{13.24 $\pm$ 3.30} & 12.41 $\pm$ 8.97 & 99.80 \\
JAX & No & 12.62 $\pm$ 5.79 & 37.27 $\pm$ 17.76 & 0.34 $\pm$ 0.01 & \textbf{100.00} \\
TensorFlow & No & \textbf{8.06 $\pm$ 3.38} & 36.99 $\pm$ 17.52 & \textbf{0.22 $\pm$ 0.02} & \textbf{100.00} \\

\hline
CCD & Yes & 570.49 $\pm$ 54.95 & 496.11 $\pm$ 43.27 & 1.15 $\pm$ 0.08 & 0.80 \\
FABRIK & Yes & 439.90 $\pm$ 40.93 & 496.08 $\pm$ 43.69 & 0.89 $\pm$ 0.05 & 0.80 \\
IPOPT & Yes & 2759.28 $\pm$ 8045.75 & \textbf{44.52 $\pm$ 88.44} & 45.31 $\pm$ 17.13 & 96.50 \\
JAX & Yes & 86.18 $\pm$ 58.17 & 67.09 $\pm$ 45.98 & 1.29 $\pm$ 0.04 & \textbf{99.60} \\
TensorFlow & Yes & \textbf{43.55 $\pm$ 30.03} & 67.40 $\pm$ 46.00 & \textbf{0.65 $\pm$ 0.10} & \textbf{99.60} \\

\hline
\end{tabular}
\caption{Performance metrics for all four tested implementations both with the simple end effector target and the complex objective $J(\boldsymbol{\theta})$. In all fields we report the mean value with the respective standard deviation. For solving time, iterations, and time per iteration, lower is better. For success rate, higher is better.}
\label{tab:performance_metrics}
\end{table*}

To evaluate the performance of our proposed TensorFlow-based inverse kinematics solver, we performed objective comparisons against several established methods: Cyclic Coordinate Descent (CCD) \cite{canutescu_cyclic_2003}, FABRIK \cite{aristidou2011fabrik}, Interior Point OPTimizer (IPOPT) \cite{wachter2002interior, wachter2006implementation}, and an implementation of our library in JAX with just-in-time (JIT) compilation. This JAX implementation is exactly the same as the TensorFlow version but replaces all TensorFlow calls with JAX calls. There are no other implementation differences and the resulting value outputs (apart from slight rounding differences) are exactly the same, given the same input seed. We also implemented our library in PyTorch using TorchScript. However, despite our best efforts, the code performed excessive recompilations during each iteration resulting in execution times of 10-20 seconds per target. For clarity, we therefore excluded the PyTorch implementation from further analysis.

Our experiments were conducted using the SMPLX skeleton \cite{SMPL-X:2019}, a widely adopted standard for human body armatures in applications that require modeling complex joint behavior. In particular, we focused on the upper extremities, testing the bones \textit{<hand>\_collar}, \textit{<hand>\_shoulder}, \textit{<hand>\_elbow} and \textit{<hand>\_wrist} for both the left and right hand. This subset was chosen because of the challenging rotational dynamics at the wrist and the strict boundary conditions at the elbow.

Each method was evaluated using two types of IK targets. The first was a simple target end-effector position where the goal was to reach a certain target position with the index finger of the respective hand, while the second was a combination of the previously described objective functions $J(\boldsymbol{\theta})$ designed to encapsulate realistic human bone boundary constraints. In all experiments, the weight of $\lambda_{\text{dist}}$ and $\lambda_{\text{known}}$ was set to 1.0, while the weight of $\lambda_{\text{look}}$ was set to 0.1. 

For consistency, each implementation was allowed a maximum of 500 iterations per run. Each run was concluded and reported as a success if the solving algorithm reached a total loss threshold of 0.005 or lower. If the implementation did not reach the threshold inside the 500 iterations limit, the run was considered a failure.

The experimental dataset was generated by first sampling one million target configurations around the SMPLX skeleton. We then filtered these to retain only those targets that could be successfully solved by any of the tested implementations, resulting in 24,531 valid configurations. To ensure a representative and evenly distributed set, we performed k-means clustering on the solved targets and selected 1010 centroids. Furthermore, to mitigate any initialization effects, measurements were taken only after the first ten targets had been processed, leaving us with 1000 sampled target locations. After each target, the SMPLX skeleton was reset to its original resting position, to not skew the iteration count between targets.

Performance metrics were computed by averaging the results over five independent runs per algorithm to account for any processing variations.

As both the JAX and TensorFlow implementation can be run with or without the GPU, we first measured the difference in iteration time for all targets with both the simple end effector and the objective function $J(\boldsymbol{\theta})$. Table \ref{tab:cpu_gpu} shows the average result over all 1000 targets. Here, both CPU configurations are faster than their GPU counterpart. Therefore, we use the CPU configuration of the JAX and TensorFlow solver for all of our performance metrics.

Table \ref{tab:performance_metrics} summarizes the performance metrics for the tested algorithms under both objective formulations. Under the simpler IK objective (without the custom objective $J(\boldsymbol{\theta})$), the TensorFlow-based solver achieved a mean solution time of 8.06 ms with an average of 36.99 iterations and a success rate of 100\%. The JAX implementation was competitive in iteration count (37.27 ± 17.76) and success rate (100\%), although its solving time was slightly higher at 12.62 ms. In comparison, IPOPT had a notably lower iteration count of 13.24 ± 3.30, indicating rapid convergence in terms of iterations, but with a higher mean solving time of 164.73 ± 132.25 ms. Conversely, both CCD and FABRIK exhibited substantially lower performance under the simple end-effector objective; while CCD achieved a moderate solving time of 25.44 ms with a success rate of 91.19\%, FABRIK performed noticeably worse, with a mean solving time of 85.28 ms.

When the custom objective function $J(\boldsymbol{\theta})$ was incorporated to simulate realistic human joint constraints, the performance gap between the methods became even more pronounced. Both CCD and FABRIK experienced substantial increases in computational cost, with mean solving times of 570.49 ms and 439.90 ms, respectively, and iteration counts approaching the maximum limit of 500 iterations per target; their success rates plummeted to under 1\%, rendering them ineffective for applications requiring stringent adherence to biomechanical constraints. IPOPT’s performance further deteriorated under the more challenging objective, with its solving time surging to 2759.28 ± 8045.75 ms and its iteration count averaging 44.52 ± 88.44, coupled with a slightly lower success rate of 96.50\%. Both the JAX and TensorFlow implementations maintained robust performance; the JAX-based solver incurred a moderate solving time of 86.18 ms with an average of 67.09 iterations and a success rate of 99.60\%, while the TensorFlow-based solver outperformed the others with a mean solving time of 43.55 ms, an iteration count of 67.40, and a success rate of 99.60\%.

\subsection{Motion Planning Results}

\begin{table}[bht]
\centering
\begin{tabular}{c c c c}
\hline
Trajectory Points & Solving Time (ms) & Success Rate (\%) \\
\hline
0 & 8.06 $\pm$ 3.38  & 100.00 \\
5 & 17.12 $\pm$ 11.01 & 99.30 \\
10 & 21.16 $\pm$ 16.92 & 97.40 \\
\hline
\end{tabular}
\caption{Distance trajectory effector performance metrics for different numbers of points in a trajectory. In all fields we report the mean value with the respective standard deviation. For solving time lower is better. For success rate higher is better.}
\label{tab:motion_simple_objective}
\end{table}

\begin{table}[bht]
\centering
\begin{tabular}{c c c c}
\hline
Trajectory Points & Solving Time (ms) & Success Rate (\%) \\
\hline
0 & 43.55 $\pm$ 30.03 & 99.60 \\
5 & 88.88 $\pm$ 77.82 & 94.99 \\
10 & 162.83 $\pm$ 125.96 & 83.67 \\
\hline
\end{tabular}
\caption{$J_{\text{trajectory}}(\boldsymbol{\theta})$ objective performance metrics for different numbers of points in a trajectory. In all fields we report the mean value with the respective standard deviation. For solving time lower is better. For success rate higher is better.}
\label{tab:motion_theta_objective}
\end{table}

We evaluated our motion planning extension on the TensorFlow implementation by comparing two configurations: a simple distance-based objective applied to the trajectory end effector $J_{\text{distance}}(\boldsymbol{\theta}) + \sum_{n=1}^{3} J_{\text{smoothness}}^{(n)}(\boldsymbol{\theta})$, and the full trajectory objective \(J_{\text{trajectory}}(\boldsymbol{\theta})\), which incorporates additional task-specific constraints and smoothness terms. We used the same weights for all hyperparameters as in the original experiment, except for the new parameter $\lambda_n$, which we set to 0.01. For both cases, we varied the number of trajectory points by testing no intermediate trajectory points, 5 trajectory points, and 10 trajectory points to assess the impact on solution time and success rate.

Table \ref{tab:motion_simple_objective} shows the performance of the distance-based objective as the number of trajectory points increases. With no intermediate points (i.e., a single-target configuration), the solver achieved an average solving time of 8.06 ms, a mean iteration count of 36.99 $\pm$ 17.52, and a success rate of 100\%. When the trajectory was extended to 5 trajectory points, the solving time increased to 17.12 ms, the iteration count increased to 83.01 $\pm$ 57.81, while the success rate decreased slightly to 99.30\%. With 10 points, the average solution time reached 21.16 ms, the iteration count increased to 102.65 $\pm$ 87.65, and the success rate lowered to 97.40\%.
Similarly, solving times under the full trajectory objective $J_{\text{trajectory}}(\mathbf{\theta})$ increased linearly with the number of trajectory points (see table \ref{tab:motion_theta_objective}). For the baseline case with no intermediate trajectory points, the solver recorded an average solving time of 43.55 ms, an average iteration count of 67.40 $\pm$ 46.00, and a success rate of 99.60\%. With 5 intermediate points, the solving time nearly doubled to 88.88 ms, the iteration count increased to 128.19 $\pm$ 113.24, and the success rate dropped to 94.99\%. When the number of trajectory points was increased to 10, the mean solving time increased to 162.83 ms, the average iteration count increased to 229.62 $\pm$ 178.41, and the success rate further decreased to 83.67\%. This strong decrease in success rate can be partially explained by the strong increase in iteration count, as 71 of our samples exceeded the maximum iteration count of 500. Doubling the iteration count could therefore lead to improved success rates for higher numbers of intermediate trajectory points. Regardless, these results indicate that the solver remains highly robust and works well within real-time constraints even as the problem dimensionality increases, especially if only a small amount of objective functions is used.

\section{Discussion}

Our experimental results demonstrate significant advantages of our TensorFlow-based inverse kinematics (IK) solver, particularly in terms of adaptability, computational speed, and reliability for complex systems with high degrees of freedom. Cyclic Coordinate Descent (CCD) methods, while efficient in simpler contexts, often encounter problems such as error accumulation and local minima when scaled up to more complex scenarios, especially under joint angle constraints \cite{xu2023combined, Boulic2007InverseKA,Kenwright2012InverseK}. These limitations are particularly pronounced in human motion modeling, which is characterized by intricate joint interdependencies and stringent biomechanical constraints. Similarly, FABRIK often exhibits inefficiencies and lower success rates when realistic boundary conditions and multiple concurrent constraints are enforced \cite{Aristidou2016ExtendingFW, martin2018natural, Kenwright2012InverseK, tao2021Extending}. 

IPOPT exhibited notable strengths in iteration efficiency, rapidly converging within fewer steps compared to all other implementations. This efficiency highlights IPOPT's effectiveness in navigating the optimization landscape quickly due to its interior-point method design, making it suitable for problems where the iteration cost is very high. However, IPOPT's computational overhead per iteration, presumably due to the high cost of computing Newtowns direction \cite{lin2021admm}, proved to be a substantial drawback limiting its practicality in real-time applications or scenarios requiring high-frequency updates. Additionally, IPOPT demonstrated sensitivity to the complexity introduced by biomechanical constraints encoded in the custom objective function, leading to increased variability and occasional convergence issues, which impacted its overall reliability compared to methods specifically optimized for computational efficiency like our TensorFlow solver.

In computer animation and gaming, character rigs typically have many degrees of freedom and require highly responsive IK solutions to achieve believable and realistic motion \cite{kenwright2022real}. Our proposed solver effectively meets these interactive performance requirements without sacrificing realism. The modular design allows seamless integration of constraints such as contact points, gaze direction, or stylistic elements without requiring retraining or modification of the underlying IK algorithms. Particularly in virtual reality and telepresence applications, where sparse tracking data points are typically available, our solver could reliably infer complete full-body poses while respecting essential human kinematic constraints such as realistic elbow extension and knee flexion limits \cite{parger2018human}.

In humanoid robotics and virtual human simulation, it could generate complex movements on demand by optimizing joint motions to achieve goals such as balance, reaching, and obstacle avoidance. This on-demand optimization accelerates development and prototyping, as one of the major benefits of our differentiable solver is its extensibility via modular objective functions. In contrast to analytic or Jacobian pseudoinverse methods that struggle to handle multiple objectives simultaneously \cite{robotics11020044}, our approach allows arbitrary constraints to be added as terms in the loss function. For instance, one can impose joint limit penalties, collision avoidance terms, or even soft biomechanical targets simply by including the corresponding differentiable cost functions in the optimization. The solver would then naturally balance all these objectives via the gradient descent process. This capability is akin to the penalty-based formulations in traditional IK solvers \cite{beeson2015trac}, but implemented in a powerful unified end-to-end differentiable manner within TensorFlow. Consequently, our IK solver easily accommodates diverse tasks. For instance, achieving precise reaching motions while preserving human-like postures can be realized by combining an end-effector position loss with a reference pose similarity term. Adjustments to motion objectives require no additional retraining or specialized algorithmic modifications to the objective function itself. Additionally, because our method fundamentally relies on the provided kinematic model, it generalizes seamlessly across different character rigs and robotic manipulators. A single solver implementation can thus manage multiple distinct avatars or robotic systems without requiring model-specific training, eliminating the overhead typically associated with neural network-based IK solutions \cite{limoyo2024generative}. Although we evaluated our IK solver on the different arm joints, this approach is fully extendable to arbitrary amounts of joints. In the future, we would like to use this approach to solve movements for the entire SMPLX skeleton and

\section{Limitations}
Despite its advantages, the differentiable inverse kinematics (IK) solver has trade-offs, notably higher computational cost per frame compared to analytical methods or pre-trained neural networks. Unlike direct feedforward network solutions, our iterative optimization approach requires more computational time, especially for highly redundant systems or scenarios requiring precise accuracy. To mitigate this, we use TensorFlow's XLA JIT compilation to optimize the computational graph, including Jacobian computations through automatic differentiation, and warm start each frame with the previous solution to minimize iterations. Although our method converges in milliseconds, it remains somewhat slower per frame than specialized analytical IK methods or trained networks, which can provide solutions in microseconds. Also, this approach inherently runs the risk of converging to local minima due to the nonconvex optimization landscape. However, by incorporating physically realistic constraints such as joint bounds and continuity from previous frames, we can significantly reduce the likelihood of unnatural poses or sudden deviations. In challenging or abruptly changing goal scenarios, the use of random restarts or secondary goals can further improve robustness, although such measures were rarely required in our testing. Furthermore, our model currently relies on an accurate kinematic model for solving the inverse kinematic problem, which is not always available. This limitation could be resolved using recent work in adaptable kinematic models, but in our work, this is not yet realized \cite{ponton2023fitted, Tejwani2023AnAR}. 

Another limitation is the naturalness of the produced motion, especially for virtual agents. Although we already have objective functions that minimize the changes in velocity, acceleration, and jerk, this can lead to sudden changes or movements that do not feel natural to an outside observer. In the future, we could implement different learned objective functions that expand upon this problem by, for example, introducing GAN discriminators for natural human motion \citet{Men2021GAN-based}, learned motion classificators from large datasets \cite{liu2024emage}, or even motion retrieval systems that align the solving with human input data, similar to \citet{mughal2024raggesture}.

\section{Conclusion}

In this paper, we have presented a TensorFlow-based inverse kinematics solver that uses automatic differentiation and just-in-time compilation to solve complex, multi-constrained IK problems efficiently and accurately. By formulating both forward and inverse kinematics as differentiable functions, our approach addresses the challenges inherent in high-degree-of-freedom systems, such as error accumulation along kinematic chains and complicated joint dynamics. Our experimental evaluation has shown that the TensorFlow-based solver consistently outperforms established inverse kinematic methods. For both simple and custom IK targets, the proposed solver achieved rapid convergence with minimal iteration times and near-perfect success rates. In contrast, conventional methods such as CCD and FABRIK struggled under realistic boundary conditions, exhibiting significantly higher iteration counts and lower success rates. The performance of our JAX-based implementation, while competitive in simpler scenarios, could not match the efficiency of the TensorFlow approach under more demanding conditions.

These results underscore the practical advantages of modern differential approaches for real-time applications in robotics, computer graphics, and biomechanics. The flexibility to define arbitrary objective functions further broadens the solver's applicability in modeling complex joint rotations and enforcing realistic constraints. Future work may explore faster optimization and extension to additional kinematic models, solidifying the role of differentiable programming in advancing inverse kinematics solutions for complex articulated systems.

\clearpage
\bibliographystyle{ACM-Reference-Format}
\bibliography{references}

\end{document}